\documentclass[runningheads]{llncs}
\usepackage{graphicx}
\usepackage{comment}
\usepackage{color, colortbl}
\definecolor{Gray}{gray}{0.9}
\usepackage{amsmath,amssymb}
\usepackage{color}
\usepackage{booktabs}

\begin{document}

\pagestyle{headings}
\mainmatter

\title{ArTIST: Autoregressive Trajectory Inpainting and Scoring for Tracking} % Replace with your title

\titlerunning{ArTIST: Autoregressive Trajectory Inpainting and Scoring for Tracking} 
\authorrunning{Fatemeh Saleh, Sadegh Aliakbarian, Mathieu Salzmann, Stephen Gould} 
\author{
Fatemeh~Saleh\inst{1,2},~Sadegh~Aliakbarian\inst{1,2},~Mathieu~Salzmann\inst{3},~Stephen~Gould\inst{1,2}
}
\institute{$^1$Australian National University, $^2$Australian Centre for Robotic Vision\\
{\tt \small \{fatemehsadat.saleh,sadegh.aliakbarian,stephen.gould\}@anu.edu.au}\\
$^3$CVLab, EPFL\\
{\tt \small mathieu.salzmann@epfl.ch}
}

\maketitle

\begin{abstract}
One of the core components in online multiple object tracking (MOT) frameworks is associating new detections with existing tracklets, typically done via a scoring function. Despite the great advances in MOT, designing a reliable scoring function remains a challenge.  
In this paper, we introduce a probabilistic autoregressive generative model to score tracklet proposals by directly measuring the likelihood that a tracklet represents natural motion. One key property of our model is its ability to generate multiple likely futures of a tracklet given partial observations. This allows us to not only score tracklets but also effectively maintain existing tracklets when the detector fails to detect some objects even for a long time, e.g., due to occlusion, by sampling trajectories so as to inpaint the gaps caused by misdetection.
Our experiments demonstrate the effectiveness of our approach to scoring and inpainting tracklets on several MOT benchmark datasets. We additionally show the generality of our generative model by using it to produce future representations in the challenging task of human motion prediction.

% \keywords{Multiple object tracking, generative models, human motion prediction}
\end{abstract}

\begin{figure*}[!h]
\vspace{-10pt}
    \centering
    \includegraphics[width=\textwidth]{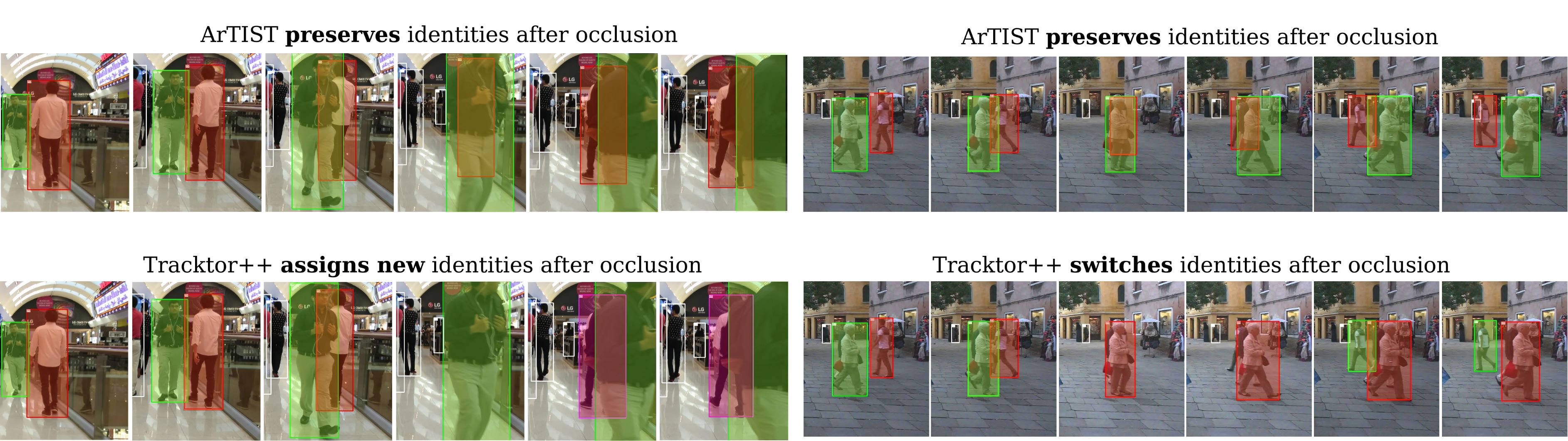}
    \caption{\textbf{Qualitative comparison of ArTIST (first row) and the SOTA Tracktor++~\cite{bergmann2019tracking} (second row).} These results evidence the effectiveness of our inpainting and scoring strategies at handling occlusions in complex and crowded scenes. Note that although Tracktor++ makes use of a person re-identification model trained on the MOT challenge, it failed to preserve the identities during the tracking process and assigned new identity (left) or switched identities (right) after an occlusion.}
    \label{fig:qualitative_intr}
    \vspace{-10pt}
\end{figure*}{}

\section{Introduction}
\label{sec:introduction}
Tracking multiple objects in a video is key to the success of many computer vision applications, such as sport analysis, autonomous driving, robot navigation, and visual surveillance. 
With the recent progress in object detection, tracking-by-detection~\cite{andriluka2008people} has become the de facto approach to multiple object tracking; it consists of first detecting the objects in the individual frames and then associating these detections with trajectories, known as tracklets. In this context, existing tracking systems can be roughly grouped into online ones~\cite{hong2016online,leal2016learning,milan2017online,maksai2017non,chu2017online,sadeghian2017tracking,zhu2018online,long2018real,yoon2018online,chu2019online,bergmann2019tracking,chu2019famnet,xu2019spatial}, where the tracklets are grown at each time step, and batch-based (a.k.a. offline) ones~\cite{tang2015subgraph,kim2015multiple,choi2015near,tang2017multiple,son2017multi,chen2017enhancing,kim2018multi,yoon2018multiple,maksai2019eliminating}, where the tracklets are computed after processing the entire sequence, usually in a multiple hypothesis tracking (MHT)~\cite{blackman2004multiple,kim2015multiple} framework. 
In this paper, we develop an online tracking system.

At the heart of most tracking-by-detection frameworks lies a scoring function aimed at assessing the quality of a tracklet after assigning it a new detection. Perhaps the most common source of information to define such a scoring function is appearance. For instance, inspired by person re-identification methods~\cite{hermans2017defense}, multiple object tracking algorithms~\cite{breitenstein2010online,yang2012online,zhang2013structure,chu2017online,chu2019online,leal2016learning} that rely on single object trackers~\cite{zhai2018deep,zhang2018robust,zhang2018learning,song2018vital,bertinetto2016staple} typically design scoring functions relying on the distance between the detections' appearance over time. Appearance, however, can be less reliable in multiple object tracking scenarios, not only because pose changes and occlusions may affect it significantly, but also because multiple targets may look very similar, for example in team sports. Furthermore, such person re-identification modules require additional training and was shown to be highly dependent on the target domain~\cite{fu2019self,deng2018image,fan2018unsupervised}.
As a consequence, many methods~\cite{milan2017online,bewley2016simple,yoon2019data,leal2015motchallenge,hamid2015joint,milan2013continuous,yoon2015bayesian,sanchez2016online,pirsiavash2011globally,dicle2013way} rather exploit geometric information, which does not suffer from these limitations. To improve robustness, recent work~\cite{sadeghian2017tracking,maksai2019eliminating,kim2018multi,fang2018recurrent} has focused on combining appearance with geometric and social information to learn scoring functions using 
recurrent neural networks (RNNs).
As acknowledged in their respective papers, while effective, training the resulting models requires a significant amount of manual data preparation, such as creating a dataset to train a good-versus-bad binary tracklet classifier~\cite{kim2018multi} or carefully balancing the data~\cite{maksai2019eliminating}, and elaborate training procedures. 

Unlike previous approaches, in this paper, we propose to learn to score tracklets directly from the tracking data at no additional data preparation cost. To this end, we design a probabilistic autoregressive model that explicitly learns the distribution of natural tracklets. This allows us to estimate the likelihood of a tracklet given only a sequence of bounding box locations. As such, we can not only compute the quality of a tracklet after assigning it a new detection, but also inpaint a tracklet missing several detections by sampling from the learned distribution. To the best of our knowledge, our approach constitutes the first attempt at filling in the gaps due to detector failures. Doing so by sampling from the distribution of natural human trajectory given an observed partial trajectory leads to a natural inpainting of the missing detections.

To summarize, our contributions are as follows:
    \textbf{(1)} We introduce a probabilistic autoregressive generative model capable of reliably scoring  a tracklet by directly measuring the likelihood that it represents natural motion.
    \textbf{(2)} Since our model learns the distribution of natural human motion, it is capable of generating multiple plausible future representations of the tracklets, and of inpainting tracklets containing missed detections.
    \textbf{(3)} We show that our geometry-based scoring function generalizes beyond the dataset it was trained on, allowing us to deploy it in diverse situations, even when the new domain differs significantly from the training one. This is due to the fact that our scoring function effectively learns the distribution of natural motions, without depending on appearance, camera viewpoint, or specific tracking metrics. 
   \textbf{(4)} Following the recent trends in MOT~\cite{bergmann2019tracking,xu2019deepmot}, we additionally demonstrate the effectiveness of our probabilistic scoring function and tracklet inpainting scheme when used in conjunction with the bounding box refinement head of~\cite{bergmann2019tracking}, which allows us to outperform the state of the art.
    \textbf{(5)} Finally, we evaluate our model's ability to generate plausible future representations in the challenging task of human motion prediction, that is, forecasting future 3D human poses given a sequence of observed ones.

Our model, named \textbf{ArTIST}, for \textbf{A}uto\textbf{r}egressive \textbf{T}rajectory \textbf{I}npainting and \textbf{S}coring for \textbf{T}racking, has a simple design and is trained with a simple negative log-likelihood loss function. 

\section{Related Work}
\label{sec:related_work}
In this section, we focus on the previous work tackling the task of multiple object tracking. For a brief review of existing approaches to human motion prediction, we refer the reader to the Appendix.

Multiple object tracking has a longstanding history in computer vision. Following the general trend in the field, most recent tracking systems follow a deep learning formalism~\cite{chen2017online,kim2018multi,sun2019deep,zhu2018online,chu2019famnet,bergmann2019tracking,xu2019spatial,maksai2019eliminating,son2017multi,sadeghian2017tracking,milan2017online,ran2019robust,liang2018lstm,wan2018online,fang2018recurrent}. Among them, closest to our approach are the ones that use recurrent neural networks,
which we thus focus on here.
The earliest RNN-based tracking framework~\cite{milan2017online} aimed to mimic the behavior of a Bayesian filter. To this end, one RNN was used to model motion and another to compute association vectors between tracklets and new detections. 
Following the success of~\cite{milan2017online} at modeling the motion with RNNs, several recurrent approaches have been proposed for MOT. In~\cite{sadeghian2017tracking}, three LSTMs were used to model the temporal dependencies between the appearance, motion, and interactions of tracklets. In the absence of occlusions, a single object tracker was used to track the different objects in the scene. To handle occlusions, this single object tracker was replaced by a Hungarian algorithm~\cite{munkres1957algorithms} based on the scores/cost matrix computed by the LSTMs to assign the detections to the tracklets. Similarly, in~\cite{ran2019robust}, a three-stream LSTM-based network was introduced to combine pose, appearance and motion information. 
In~\cite{liang2018lstm}, a Siamese LSTM was used to model the position and velocity of objects in a scene 
for scoring and assignment purposes. 
In~\cite{wan2018online}, a Siamese LSTM on motion and appearance was employed to provide scores to a Hungarian algorithm that merged short tracklets, initially obtained with a Kalman filter. In~\cite{fang2018recurrent}, two recurrent networks were used to maintain external and internal memories for modeling motion and appearance features to compute the scores to be used in the assignment process. 

While the previous algorithms worked in an online manner, recurrent models have also been used in offline tracking pipelines. For instance, in~\cite{maksai2019eliminating}, LSTMs were used to score the tracklets in an MHT framework. To this end, a recurrent scoring function that utilizes appearance, motion, and social information was trained to optimize a proxy of the IDF1 score~\cite{ristani2016performance}. While this achieved promising performance, as acknowledged by the authors, it required manual parameter tuning, data augmentation, and carefully designing the training procedure.
In~\cite{kim2018multi}, LSTMs were used to decide when to prune a branch in an MHT framework. This approach, called bilinear LSTMs, used a modified LSTM cell that takes as input the appearance and motion. However, the appearance-based and motion-based models were first pre-trained separately. The way the LSTM cells were modified to handle appearance information when learning longer-range dependencies was shown to be sensitive to the quality of the detections.

In general, most of the top-performing approaches use appearance information~\cite{bergmann2019tracking,ristani2018features,chu2019famnet,xu2019spatial,maksai2019eliminating,sadeghian2017tracking,kim2018multi,zhu2018online,long2018real}. However, to get the best out of appearance, one needs to re-train/fine-tune the appearance model on each target dataset. This limits the applicability of these approaches to new datasets. Moreover, in datasets such as MOT17~\cite{milan2016mot16}, all the test sequences have a similar counterpart in the training sequences, which significantly simplifies the appearance-based models' task, but does not reflect reality. Unlike these approaches, ArTIST only leverages geometric information for training, without  depending on the appearance of the target dataset. In fact, as will be demonstrated in our experiments, ArTIST does not even need to see the geometric information of the target dataset as it only uses this information to learn a distribution over natural human motion, which can be achieved with any MOT dataset covering sufficiently diverse scenarios, such as moving/static cameras, different camera viewpoints, and crowded scenes. Furthermore, in contrast to prior methods that use multiple streams to handle different modalities~\cite{sadeghian2017tracking,fang2018recurrent,kim2018multi,ran2019robust}, manipulate the training data~\cite{kim2018multi}, or design data-sensitive and complex loss functions~\cite{maksai2019eliminating}, our model relies on a very simple recurrent network architecture with a simple negative log-likelihood loss function and can be trained directly on any tracking dataset without any data manipulation or augmentation.

Note that a number of methods, such as Social LSTM~\cite{alahi2016social} and Social GAN~\cite{gupta2018social}, utilize generative models to encode the social behavior of crowd motion. Since they focus on modeling social information, these approaches are in general not comparable to MOT ones and thus go beyond the scope of this work.

\section{Proposed Method}
\label{sec:proposed_method}
We address the problem of online tracking of multiple objects in a scene. Our approach relies on two main steps in each time-frame: Scoring how well a detection fits in an existing tracklet and assigning the detections to the tracklets. Below, we first describe our overall tracking pipeline. We then delve into the details of our scoring function and assignment strategy.

\subsection{Multiple Object Tracking Pipeline}
\label{subsec:MOT_pipeline}
As in many other online tracking systems, we follow a  tracking-by-detection paradigm~\cite{andriluka2008people}. 
Let us consider a video of $T$ frames, where, for each frame, we are provided with a set of detections computed by, e.g., Faster-RCNN~\cite{girshick2015fast}, DPM~\cite{felzenszwalb2009object}, or SDP~\cite{yang2016exploit}. This yields an overall detection set for the entire video denoted by $\mathcal{D}^{1:T} = \{D^1, D^2, ..., D^T\}$, where $D^t=\{d_1^t, ..., d_n^t\}$ is the set of all detections\footnote{While in practice the number $n$ varies across the different times $t$, we ignore this dependency here to simplify notation.} at time $t$, with $d_i^t\in\mathbb{R}^4$, i.e., the 2D coordinates $(x,y)$ of the top-left bounding box corner, its width $w$ and height $h$. We initialize a first set of tracklets $\mathcal{T}$ with the detections in the first frame $\mathcal{D}^1=\{d_1^1, ..., d_n^1\}$. From the second time-step to the end of the video, the goal is to expand the tracklets by assigning the new detections to their corresponding tracklets. 
Throughout the video, new tracklets may be created, and appended to the set of tracklets $\mathcal{T}$, and existing tracklets may be terminated, and removed from $\mathcal{T}$. 

To grow a tracklet $\mathcal{T}_j$, at time $t$,
we compute a tracklet proposal $\hat{\mathcal{T}}_j^i$ for each new detection, by appending the detection to $\mathcal{T}_j$, and compute the likelihood of each proposal under our scoring model. We compute such likelihoods for all tracklets in $\mathcal{T}$ at time $t$, and then assign the detections to the tracklets by solving a linear program using the Hungarian algorithm~\cite{munkres1957algorithms}. As a result of  this linear assignment, some detections will be assigned to some tracklets. Other detections may not be assigned to any tracklet, and thus \textit{may} serve as starting point for new tracklets. Conversely, some tracklets may not be assigned any detection, which \textit{may} lead to their termination if they remain unassigned for certain period.

Given this MOT pipeline, in the remainder of this section, we describe the ArTIST architecture that allows us to score each tracklet proposal and inpaint a tracklet when the detector fails due to, e.g., occlusion or motion blur.

\subsection{ArTIST Architecture}

\begin{figure}[t]
    \centering
    \includegraphics[width=\textwidth]{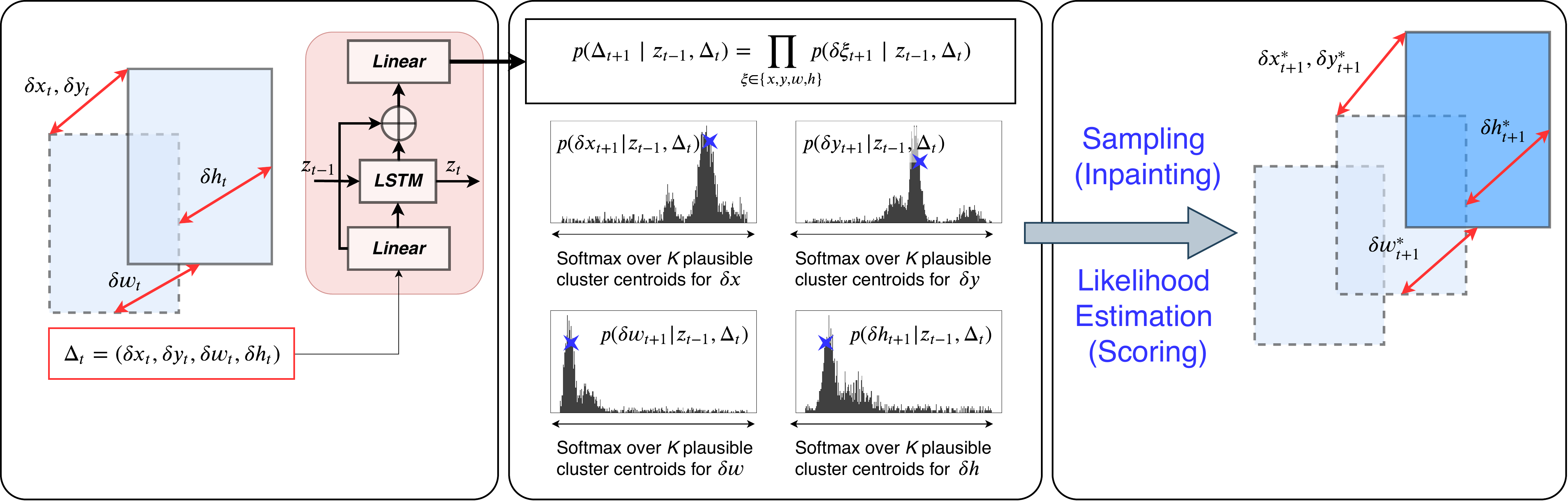}
    \caption{\textbf{ArTIST framework.} \textbf{(Left)} ArTIST relies on a recurrent residual architecture to represent motion velocities. \textbf{(Middle)} Given the motion velocity representation at time $t$, ArTIST estimates a distribution over the next (i.e., at time $t+1$) bounding box location. \textbf{(Right)} Given the estimated distributions, one can either generate a new bounding box velocity ($\delta x^*_{t+1}, \delta y^*_{t+1}, \delta w^*_{t+1}, \delta h^*_{t+1}$) by sampling (indicated by blue stars over distributions) from the distributions, or evaluate the likelihood of an observed bounding box under the model. 
    }
    \label{fig:LSTM_architecture}
\end{figure}

\label{subsec:artist_arch}
ArTIST is a probabilistic autoregressive generative model that aims to explicitly learn the distribution of natural tracklets. As an estimator, ArTIST is capable of determining the likelihood of each tracklet. As a generative model, ArTIST is capable of generating multiple plausible continuations of a tracklet by multinomial sampling from the estimated distribution at each time-step. 

The probability of a tracklet $\mathcal{T}_j$ in an autoregressive framework is defined as 
\begin{align}
    p(\mathcal{T}_j) = p(b_1)\prod_{t=2}^{T}p(b_t \mid b_{t-1}, b_{t-2}, ..., b_1)\;,
    \label{eq:autoregressive}
\end{align}
where $b_t$ is the bounding box representation assigned to $\mathcal{T}_j$ at time $t$.\footnote{Note that, for simplicity, we ignore the index $j$ in $b_t$.} To model this, because each bounding box is represented by its position, which is a continuous variable, one could think of learning to regress the position in the next frame given the previous positions. However, regression does not explicitly provide a distribution over natural tracklets. Furthermore, regression can only generate a single deterministic continuation of a tracklet, which does not reflect the stochastic nature of, e.g., human motion, for which multiple continuations may be equally likely. 

To remedy this, inspired by PixelRNN~\cite{oord2016pixel}, we propose to discretize the bounding box position space. This allows us to model $p(\mathcal{T})$ as a discrete distribution, with every conditional distribution in Eq.~\ref{eq:autoregressive} modeled as a multinomial distribution with a \textit{softmax} layer. However, unlike PixelRNN-like generative models that discretize the space by data-independent quantization, e.g., through binning, we propose to define a data-dependent set of discrete values by clustering the motion velocities, i.e., $\delta x$, $\delta y$, $\delta w$, and $\delta h$ between consecutive frames, normalized by the width and height of the corresponding frames. This makes our output space shift and scale invariant. We then define the discrete motion classes as the cluster centroids. In practice, we use the non-parametric k-means clustering algorithm to obtain $K$ clusters. 

Our ArTIST architecture is depicted by Fig.~\ref{fig:LSTM_architecture}.
ArTIST relies on a recurrent residual architecture to represent motion velocities. At each time-step $t$, it takes as input a motion velocity represented by $\Delta_t = (\delta x_t, \delta y_t, \delta w_t, \delta h_t)$. Given this input and the hidden state computed in the last time-step $z_{t-1}$, it then predicts a distribution over the motion velocity for time $t+1$, i.e., $p(\Delta_{t+1}\mid z_{t-1}, \Delta_t)$. This matches the definition in Eq.~\ref{eq:autoregressive}, since $z_{t}$ carries information about all previous time-steps.

Training ArTIST requires only the availability of a tracking dataset, and we exploit individual tracklets, without simultaneously considering multiple tracklets in the scene. Each tracklet in the dataset is defined by a sequence of bounding boxes, normalized by the width and height of the corresponding frames, from which we extract velocities $\{\Delta_\cdot\}$. Since we only aim to estimate a probability distribution over the bounding box position in the next time-step, we train our model with a simple negative log-likelihood loss function.
From the estimated distribution one can either measure the likelihood of a tracklet given a detection bounding box, or inpaint a tracklet to fill in a gap caused by missing detections. We discuss these two scenarios below.

\begin{figure*}[t]
    \centering
    \includegraphics[width=\textwidth]{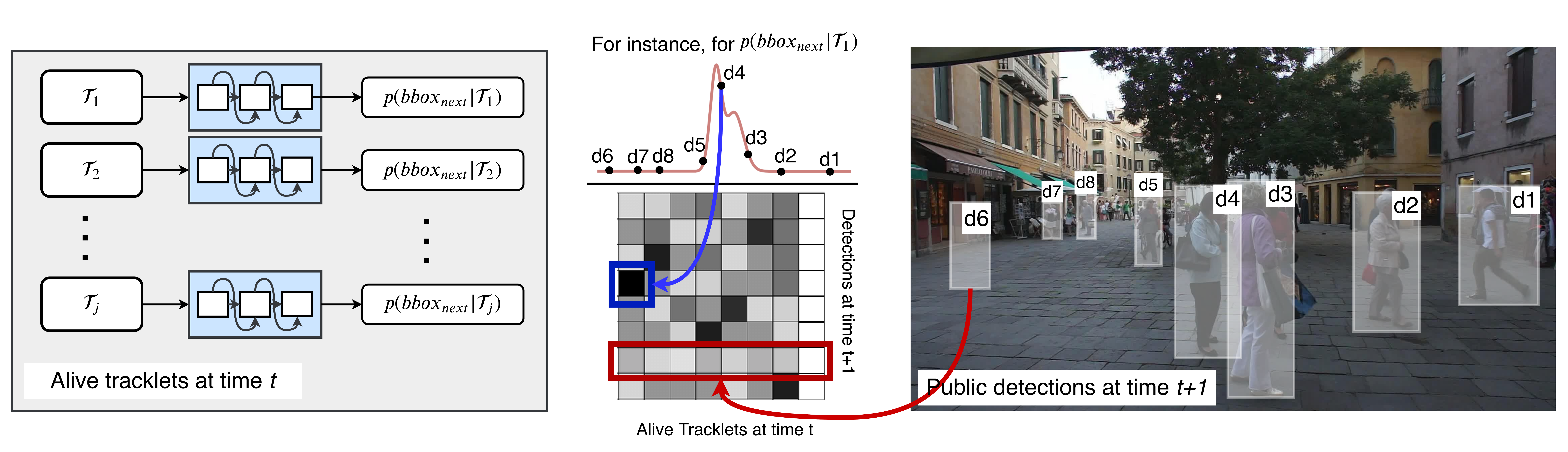}
    \caption{\textbf{Scoring tracklets with ArTIST.} To assign each of the detections provided at time $t+1$ (right image) to a  tracklet, we compute the probability distribution of the next bounding box for each tracklet $\mathcal{T}_1, ..., \mathcal{T}_j$, as shown on the left. Then, for each tracklet, we compute the negative log-likelihood of every detection, and take the negative log-likelihood of a detection $d_i$ under the model for tracklet $\mathcal{T}_j$ as the cost of assigning $d_i$ to $\mathcal{T}_j$. The Hungarian algorithm then takes the resulting cost matrix (middle) as input and returns an assignment. Here, the blue box indicates that $d_4$ is the best match for $\mathcal{T}_1$, resulting in the lowest assignment cost. The red box shows that $d_6$, which is a false detection, results in high assignment costs for all tracklets.}
    \label{fig:artist}
\end{figure*}

\paragraph{\textbf{Tracklet Scoring.}}
To score how likely a detection at time $t$ is to be the continuation of a tracklet, we input the sequence of motion velocities corresponding to the existing tracklet to ArTIST as shown in Fig.~\ref{fig:artist}. 
Given this sequence, the model then estimates a probability distribution over the location of the bounding box at time $t$. We then take the likelihood of the observed detection given the estimated distribution as a score for the tracklet-detection pair. Specifically, we compute the $\Delta$ for any detection with respect to the previous observation (or inpainted bounding box if the previous time-step was inpainted). We then take the probability estimated for the centroid closest to this $\Delta$ as likelihood. In practice, we assume independence of the bounding box parameters, i.e., $\delta x$, $\delta y$, $\delta w$, and $\delta h$. Therefore, we have four sets of clusters and thus four probability distributions estimated at each time-step. We then compute the likelihood of a bounding box as the product of the probabilities of the components, that is,
\begin{align}
    p(\Delta_{t+1} \mid z_{t-1}, \Delta_t) &= \!\!\!\!\prod_{\xi \in \{x, y, w, h\}} \!\!\!\! p(\delta \xi_{t+1} \mid z_{t-1}, \Delta_t).
\end{align}
In practice, we do this in log space, summing over the log of the probabilities.

\paragraph{\textbf{Tracklet Inpainting.}}
In the real world, detection failures for a few frames are quite common due to, e.g., occlusion or motion blur. Such failures complicate the association of future detections with the tracklets, and thus may lead to erroneous tracklet terminations. Our approach overcomes this by inpainting the tracklets for which no detections are available. Let us consider the scenario where a tracklet was not assigned any detection in the past few frames. We now seek to check whether a new detection at the current time-step belongs to it. For our model to compute a likelihood for the new observation, we need to have access to the full bounding box sequence up to the previous time-step. To this end, we use our model to inpaint the missing observations.
Specifically, since ArTIST estimates a distribution over the future bounding box position at every time, if no bounding box is assigned to a tracklet at a given time, we can sample one from the  distribution estimated at the previous time-step. Sampling can in fact be done in a recursively to create a full sequence of observations and inpaint the missing bounding boxes, which, in turn, allows us to score a new detection. 

To account for the fact that motion is stochastic by nature, especially for humans, we sample $\mathcal{S}$ candidates for the whole subsequence to inpaint from the estimated distribution and get multiple plausible inpainted tracklets. Since ArTIST relies solely on geometric information, on its own, it cannot estimate which of the $\mathcal{S}$ inpainted options are valid. To do so, we use a tracklet rejection scheme, which allows us to make a more reliable decision that an inpainted tracklet does not deviate too much from its actual direction. We elaborate on our tracklet rejection scheme below.

\paragraph{\textbf{Tracklet Rejection Scheme.}} As discussed above, our model is capable of inpainting the missing observations when a detector fails to detect an object for a few frames. Our model also accounts for the stochasticity of human motion, and thus generates multiple plausible candidates for inpainting. To select one of these candidates, if there is one to be selected, we compute the intersection over union (IOU) of the last generated bounding box with all the detections in the scene. The model then selects the candidate with highest IOU, if it surpasses a threshold. However, in some cases, the last generated bounding box of one of the candidates may overlap with a false detection or a detection for another object, i.e., belonging to a different tracklet. To account for these ambiguities, we continue predicting  boxes for all candidates for $t_{TRS}$ frames. We then compute the IOU  with the detections of not only the current frame, but also the $t_{TRS}$ frames ahead. ArTIST then selects the candidate with the maximum sum of IOUs. This allows us to ignore candidates matching a false detection or a detection for another object moving in a different direction. However, this may not be enough to disambiguate all cases, e.g., the detections belonging to other tracklets that are close-by and moving in the same direction. We treat these cases in our assignment strategy discussed below. Note that, in practice, we use a small $t_{TRS}$, e.g., 2 or 3 frames, and thus our approach can still be considered as online tracking. A more detailed illustration of our tracklet rejection scheme is provided in Fig. 4 in the Appendix.

\subsection{Assignment}
\label{subsec:assignment}
To assign the detections to the tracklets at each time-step, we use the linear assignment found by the Hungarian algorithm. The Hungarian method relies on a cost matrix $C$, storing the cost of assigning each detection to each tracklet. 
In our case, the costs are negative log-likelihoods computed by ArTIST. Let us denote by 
$C_{ij} = \log p(\langle d_i,\mathcal{T}_j\rangle)$ the negative log-likelihood of assigning detection $i$ to tracklet $j$.
The Hungarian algorithm then returns the indices of associated tracklet-detection pairs by solving 
$A^\star = \arg\min_{A} \sum_{i,j}C_{ij}A_{ij}$,
where $A\in[0,1]^{N\times M}$ is the assignment probability matrix, with $N$ the number of detections and $M$ the number of tracklets. 
This matrix satisfies the constraints $\sum_j{A_{ij}=1},\;\forall i$ and $\sum_i{A_{ij}=1},\;\forall j$.

In practice, to account for the fact that we are less confident about the tracklets that we inpainted, we run the Hungarian algorithm twice. First, using only the tracklets whose scores at the previous time-step were obtained using actual detections; second, using the remaining tracklets obtained by inpainting and the unassigned detections. 

\section{Experiments}

In this section, we empirically evaluate different aspects of ArTIST and compare it with existing methods. Importantly, to show the generality of our approach, we do not rely on \emph{any} target tracking datasets to train ArTIST. Instead, we use the PathTrack dataset~\cite{manen2017pathtrack}, one of the largest publicly available multiple object tracking dataset with more than 15,000 person trajectories in 720 sequences. In our experiments, bold numbers indicate the best results and underlined numbers the second best ones. We provide the implementation details of our approach in the Appendix. 

\paragraph{\textbf{Datasets.}} We use the multi-object tracking benchmarks MOTChallenge\footnote{https://motchallenge.net/} and JRDB~\cite{martin2019jrdb}. MOTChallenge consists of several challenging pedestrian tracking sequences with moving and stationary cameras capturing the scene from various viewpoints and at different frame rates. We report our results on the three benchmarks of this challenge, 2D MOT2015~\cite{leal2015motchallenge}, MOT16~\cite{milan2016mot16}, and MOT17~\cite{milan2016mot16}. MOT17 contains 7 training-testing sequence pairs with similar statistics. Three sets of public detections namely DPM~\cite{felzenszwalb2009object}, Faster R-CNN~\cite{girshick2015fast} and SDP~\cite{yang2016exploit} are provided with the benchmark. The sequences of MOT16 are similar to that of MOT17, with detections computed only via DPM. 2D MOT2015 contains 11 training and 11 testing sequences. For each testing sequence, there is a sequence with roughly similar statistics in the training data. This benchmark provides ACF~\cite{dollar2014fast} detections for all the training and testing sequences.  For all our experiments, we use the \textit{public} detections provided with the datasets. Since ArTIST relies on none of the aforementioned datasets during training, we perform ablation studies on the MOT17 training set with the public Faster-RCNN detections.
JRDB is a recent dataset collected from the social mobile manipulator JackRabbot. The dataset contains 64 minutes of multi-modal sensor data, including stereo cylindrical $360^{\circ}$ RGB video at 15 fps. These sequences were captured from traditionally underrepresented scenes, such as indoor environments and pedestrian areas, from both a stationary and navigating robot platform. 

\paragraph{{\textbf{Evaluation Metrics.}}} 
To evaluate MOT approaches, we use standard metrics~\cite{ristani2016performance,bernardin2008evaluating} of MOT Accuracy (MOTA), ID F1 Score (IDF1), the number of identity switches (IDs), mostly tracked (MT), mostly lost (ML), false positives (FP), and false negatives (FN). Details on these metrics are provided in the Appendix.

\subsection{Ablation Study}
\label{sec:ablation}
In this section, we evaluate different components of ArTIST. Specifically, we evaluate the effect of multinomial sampling, inpainting, tracklet rejection, and bounding box refinement. We further evaluate the generalization of our trajectory inpainting to the challenging task of human motion prediction.
For these experiments
we use the MOT17 training set with the public Faster-RCNN detections. For each experiment, in addition to the main two metrics, MOTA and IDF1, we provide and analyze the metrics that best convey the intuition and message of the experiment. For human motion prediction, we use the Human3.6M dataset~\cite{ionescu2013human3} with the standard error metric, i.e., the joint displacement error~\cite{mao2019learning}.

\paragraph{\textbf{Effect of Inpainting.}}
Most existing trackers aim to
reduce FP. However, they often ignore improving FN. We observed that FN can be considerably reduced if a tracker can fill in the gaps between detections to create even longer trajectories. This, in our approach, can be done by tracklet inpainting. In Table~\ref{tab:ablation}(a), we compare ArTIST with and without inpainting tracklets. As can be seen from the table, there is a direct correlation between FN and MOTA. This is because there are considerably more FN than FP, hence its effect on MOTA is much greater. 
This is important in scenarios where one needs to track objects even when they are occluded, e.g., in autonomous driving. As also acknowledged by~\cite{leal2017tracking}, in surveillance scenarios, it is typically more important to have very few FN so that no person is missed, while a few FP can be easily managed by humans in the loop.
We observe that, while the inpainted tracklets resemble natural human motion, not all inpainted boxes correctly match the ground truth, leading to an increase in FP and IDs. However, since FN is typically two to three orders of magnitude higher than FP and IDs, we see an overall improvement in tracking. Note also that ArTIST with inpainting is capable of keeping correct tracklets for longer time periods, resulting in much higher MT and lower ML. 
\begin{table}[t]
    \centering
    \caption{Ablation study on MOT17 training set with public Faster R-CNN detections.}
    \label{tab:ablation}
    \begin{tabular}{c @{ } c}
    \scalebox{0.58}{
    \begin{tabular}{l c c c c c c c}
            \multicolumn{8}{c}{\textbf{(a) Influence of tracklet inpainting)}}\\
        \toprule
    Setting  & MOTA $\uparrow$ & IDF1 $\uparrow$ & IDs $\downarrow$ & MT $\uparrow$ &  ML $\downarrow$ & FP $\downarrow$ & FN $\downarrow$ \\
    \midrule
    
    ArTIST (w/o Inp.)  & 48.0 & 54.1 & \textbf{353} & 15.4 & 39.9 & \textbf{1,200} & 56,838 \\
    ArTIST (w/ Inp.)  & \textbf{50.3} & \textbf{56.7} & 515 & \textbf{22.9} & \textbf{31.8} & 3,277 & \textbf{52,010} \\
    \bottomrule
    
    \multicolumn{8}{c}{\textbf{(b) Influence of Multinomial Sampling and tracklet rejection)}}\\
    \toprule
          Method & Sampling & TRS & MOTA $\uparrow$ & IDF1 $\uparrow$ & IDs $\downarrow$ &  MT $\uparrow$ &  ML $\downarrow$\\
         \midrule
         ArTIST & Top-1 & No
         & 50.0 & 54.1 & 529 & 21.8 & 33.7\\
         ArTIST & Multi. & No
         & 49.7 & 54.9 & 632 & \textbf{24.0} & \textbf{31.0}\\
         ArTIST & Multi. & Yes 
         & \textbf{50.3} & \textbf{56.7} & \textbf{515} & 22.9 & 31.8\\
         \bottomrule
        % \\
    \end{tabular}
    }
&
\scalebox{0.6}{
    \begin{tabular}{l c c c c c}
    \multicolumn{6}{c}{\textbf{(c) Influence of refining bounding boxes as in Tracktor~\cite{bergmann2019tracking})}}\\
    \toprule
          Setting & MOTA $\uparrow$ & IDF1 $\uparrow$ & IDs $\downarrow$ &  MT $\uparrow$ &  ML $\downarrow$\\
         \midrule
         ArTIST & 50.3 & 56.7 & 515 & 22.9 & 31.8\\
         ArTIST* & \textbf{62.4} & \textbf{68.8} & \textbf{255} & \textbf{37.5} & \textbf{22.3}\\
         \midrule
         Tracktor~\cite{bergmann2019tracking} & 61.5 & 61.1 & 1,747 & 33.5 & 20.7\\
         Tracktor+reID~\cite{bergmann2019tracking} & 61.5 & 62.8 & 921 & 33.5 & \textbf{20.7}\\
         Tracktor+CMC~\cite{bergmann2019tracking} & 61.9 & 64.1 & 458 & 35.3 & 21.4 \\
         Tracktor++(reID+CMC)~\cite{bergmann2019tracking} & 61.9 & 64.7 & 326 &35.3 & 21.4\\
        %  Tracktor++(reID+CMC)(Our replication) & 61.7 & 65.0 & 292 &35.5 & 20.8 & 694\\
         ArTIST* & \textbf{62.4} & \textbf{68.8} & \textbf{255} & \textbf{37.5} & 22.3\\
         \bottomrule
    \end{tabular}
    }
    \end{tabular}
\end{table}{}

\paragraph{\textbf{Effect of Multinomial Sampling.}} As discussed in Section~\ref{subsec:artist_arch}, ArTIST, as a generative model, can generate multiple plausible continuations of a tracklet by multinomial sampling from the estimation distributions. The more samples generated by the model, the higher the chance of finding the motion that is closest to the ground truth. In Table~\ref{tab:ablation}(b), we compare a model that ignores the stochasticity in human motion, and thus greedily generates a single continuation of a tracklet for inpainting, with one that takes stochasticity into account. Note that, with more inpainted options, the model achieves better performance. However, large numbers of samples may introduce ambiguities in the system, causing a decrease in tracking performance. To handle this, we disambiguate such scenarios using our tracklet rejection strategy, whose results are provided in the third row of Table~\ref{tab:ablation}(b). 
By this experiment, we observe that, for sequences captured by a static camera, and for tracklets with relatively long observations, Top-1 sampling performs reasonably well, almost on par with multinomial sampling. This is due to the fact that, with long motion observations, ArTIST captures the motion pattern and can reliably predict the future. However, when it comes to moving cameras or newly born tracklets (with relatively short observations), multinomial sampling (with tracklet rejection) leads to more reliable tracking.

\paragraph{\textbf{Effect of Bounding Box Refinement.}}
A number of recent tracking techniques~\cite{bergmann2019tracking,xu2019deepmot,maksai2019eliminating} refine the bounding boxes computed by the detectors. In particular,~\cite{bergmann2019tracking,xu2019deepmot} use Faster R-CNN~\cite{girshick2015fast} with ResNet-101~\cite{he2016deep} and Feature Pyramid Networks (FPN)~\cite{lin2017feature} trained on the  MOT17Det~\cite{milan2016mot16} pedestrian detection dataset to refine the public detections provided with the MOTChallenge. 
Note that, as also acknowledged by~\cite{bergmann2019tracking}, for the comparison with methods that use public detections to be fair, the new trajectories are still initialized from the public detection bounding boxes, and thus refinement is not used to detect a new bounding box. In fact, the bounding box regressor and classifier is used to obtain refined scores and bounding box coordinates, respectively. In this experiment, we evaluate the following two aspects. First, in Table~\ref{tab:ablation}(c, top), we show how the recent trend of detection refinement leads to better tracking quality in our ArTIST framework; and second, in Table~\ref{tab:ablation}(c, bottom), we compare ArTIST with refined detection, denoted by ArTIST* with the state-of-the-art approach using such a refinement strategy~\cite{bergmann2019tracking}. In particular, Tracktor~\cite{bergmann2019tracking} exploits the regression head of a detector to perform temporal realignment of person bounding boxes. Table~\ref{tab:ablation}(c, bottom) compares different settings of~\cite{bergmann2019tracking} with ArTIST*. These results clearly demonstrate the effectiveness of our probabilistic autoregressive inpainting and scoring function. The large improvements in IDF1 (+7.7\%), MOTA (+0.9\%), ID switch (-1,492), and MT (+4\%) show that our model manages to preserve identities after long occlusions and in crowded scenes. We additionally compare our approach with Tracktor++ that uses additional modules, such as Camera Motion Compensation (CMC) and person re-identification, to handle occlusions and moving cameras during tracking. Note that person re-identification is highly domain-dependent~\cite{fu2019self,deng2018image,fan2018unsupervised} and requires fine-tuning on the target domain to achieve a reasonable performance. By contrast, as also depicted by Fig.~\ref{fig:qualitative_intr}, ArTIST*, which solely relies on geometric information for inpainting and scoring, outperforms Tracktor++ in almost all MOT metrics without requiring such fine-tuning.
\begin{table}[t]
    \centering
    \caption{Human motion prediction results on Human3.6M dataset. 
    }
    \label{tab:human3.6m}
    \scriptsize
    \setlength\extrarowheight{-3pt}
    \scalebox{0.8}{
    \begin{tabular}{c @{ }@{ } c}
    Walking & Eating \\
    \begin{tabular}{l @{ }@{ } @{ }@{ }c @{ }@{ }c @{ }@{ }c @{ }@{ }c @{ }@{ }c @{ }@{ }c}
    
    %%%%% WALKING
    \toprule
         Method & 80 & 160 & 320 & 400 & 560 & 1000 \\
    \midrule
    Residual Sup~\cite{martinez2017human} & 
    23.8  &  40.4 &   62.9 &   70.9 & 73.8 & 86.7\\
    ConvSeq2Seq~\cite{li2018convolutional} & 
    27.1   & 31.2  &  53.8  &  61.5 & 59.2 & 71.3\\
     LTD GCN~\cite{mao2019learning} & 
    \textbf{8.9}  &   \textbf{15.7}  &  \textbf{29.2}  &  \textbf{33.4} &  \textbf{42.3} & \textbf{51.3} \\
    \midrule
    
    ArTIST& 
    \underline{14.0} & \underline{24.9} & \underline{36.1} & \underline{40.9} & \underline{46.6} & \underline{54.1}\\
   
    \bottomrule
    \end{tabular}
    
    &
    %%%%% EATING
        \begin{tabular}{l @{ }@{ } @{ }@{ }c @{ }@{ }c @{ }@{ }c @{ }@{ }c @{ }@{ }c @{ }@{ }c}
    \toprule
      Method &    80 & 160 & 320 & 400 & 560 & 1000 \\
    \midrule
    
        Residual Sup~\cite{martinez2017human} & 
        17.6  &  34.7  &  71.9 &   87.7 & 101.3  & 119.7\\
    ConvSeq2Seq~\cite{li2018convolutional} & 
    13.7  &  \underline{25.9} &   52.5  &  63.3 & 66.5 & 85.4\\
    LTD GCN~\cite{mao2019learning} & 
    \textbf{8.8}  &   \textbf{18.9}  &  \underline{39.4}  & \underline{ 47.2} &  \underline{56.6}  & \underline{68.6}\\
    \midrule
    
    ArTIST& \underline{16.7} & 28.8 & \textbf{37.2} & \textbf{38.8} & \textbf{47.9} &\textbf{ 68.3}\\
    \bottomrule
    \end{tabular}
    
    \\
    Smoking & Discussion \\
    \begin{tabular}{l @{ }@{ } @{ }@{ }c @{ }@{ }c @{ }@{ }c @{ }@{ }c @{ }@{ }c @{ }@{ }c}
    
    %%%%% SMOKING
    \toprule
         Method & 80 & 160 & 320 & 400 & 560 & 1000 \\
    \midrule
    Residual Sup~\cite{martinez2017human} & 
    19.7 &   36.6  &  61.8 &   73.9 & 85.0  & 118.5\\
    ConvSeq2Seq~\cite{li2018convolutional} & 
    11.1  &  21.0  &  33.4  &  38.3 & 42.0 & 67.9\\
     LTD GCN~\cite{mao2019learning} & 
    \underline{7.8}   &  \underline{14.9}  &  \underline{25.3}  &  \underline{28.7} &  \textbf{32.3}  & \underline{60.5}\\
    \midrule
    
   ArTIST& 
    \textbf{ 5.9} & \textbf{10.7} & \textbf{21.0} &\textbf{ 25.8} & \underline{34.6} & \textbf{56.0}\\
    
    \bottomrule
    \end{tabular}
    
    &
    %%%%% DISCUSSION
    \begin{tabular}{l @{ }@{ } @{ }@{ }c @{ }@{ }c @{ }@{ }c @{ }@{ }c @{ }@{ }c @{ }@{ }c}
    \toprule
         Method & 80 & 160 & 320 & 400 & 560 & 1000 \\
    \midrule
    
    Residual Sup~\cite{martinez2017human} & 
    31.7  &  61.3  &  96.0  &  103.5 & 120.7 & 147.6\\
    ConvSeq2Seq~\cite{li2018convolutional} & 
    18.9  &  39.3  &  67.7  &   75.7 & 84.1 & 116.9\\
    LTD GCN~\cite{mao2019learning} & 
    \textbf{9.8}   &  \textbf{22.1}  &  \textbf{39.6}  &   \textbf{44.1} & \textbf{70.5} & \textbf{103.5}\\
    \midrule
    
     ArTIST& 
     \underline{16.8} & \underline{35.8} & \underline{58.4} & \underline{63.9} & \underline{81.5} & \underline{108.7}\\
    \bottomrule
    \end{tabular}
    
    \end{tabular}
    }
\end{table}{}

\paragraph{\textbf{Evaluating the Generalizability of our Inpainting.}}
To further evaluate the capability of ArTIST to generate plausible future representations, we tackle the problem of 3D human motion prediction, where the 3D annotations were obtained by motion capture (MoCap). The goal here is to generate a future human motion given observations of the past motion. We evaluate our approach on Human3.6M, following the standard settings for training and evaluation~\cite{martinez2017human,li2018convolutional,mao2019learning}. Specifically, we follow the observation of the state-of-the-art human motion prediction model~\cite{mao2019learning} and train our model on 3D joint positions. Unlike tracking where we assume independence between bounding box parameters, here we consider all joints to be related to each other. Therefore, we cluster the entire pose velocities (in 96D, i.e., 32 joints in 3D) into 1024 clusters. This means that we only consider 1024 possible transitions between two frames. Although this seems under-representative of how freely humans can move between two consecutive time instants, ArTIST achieves very good performance compared to existing models, as shown in Table~\ref{tab:human3.6m}. 
Note that, although ArTIST is not designed specifically for human motion prediction (as opposed to other works~\cite{martinez2017human,li2018convolutional,mao2019learning} that designed complex architectures for this particular task), using it with virtually no modifications (except for the input representation) yields the best/second-best results on this challenging task.

\begin{table}
    \centering
     \centering
     \caption{\textbf{Results on different MOTChallenge benchmark datasets}, either in an online or offline framework. For each test set: (Top) Approaches that utilize only geometric features for tracking. (Bottom) Approaches that additionally utilize appearance information for tracking.}
    \label{tab:motchallenge}
    \scriptsize
    \tabcolsep=0.2cm
    \scalebox{0.8}{
    \begin{tabular}{c}
         %%%%%%% MOT15
         \textbf{Results on 2D MOT2015 Test Set} \\
    \begin{tabular}{ l c  c c c c c c c}
    \toprule
    Method & Mode  & MOTA $\uparrow$& IDF1 $\uparrow$ & IDs $\downarrow$ & MT $\uparrow$ &  ML $\downarrow$ & FP $\downarrow$ & FN $\downarrow$ \\
    \midrule
    
    \multicolumn{9}{c}{Geometry-based Tracking Methods}\\
    \midrule
    
    CEM~\cite{milan2013continuous} & Offline 
    & 19.3 & -  & 813 & 8.5 & 46.5 & 14,180 & 34,591\\
    
    JPDA~\cite{hamid2015joint}  & Offline 
    & 23.8  & 33.8 & 365 & 5.0 & 58.1 & 6,373 & 40,084 \\
    DP-NMS~\cite{pirsiavash2011globally} & Offline 
     & 14.5& 19.7 & 4,537 & 6.0 & 40.8 & 13,171 & 34,814\\
    LP2D~\cite{leal2015motchallenge} & Offline 
    & 19.8  & - & 1,649 & 6.7 & 41.2 & 11,580 & 36,045 \\
    SMOT~\cite{dicle2013way} & Offline 
    & 18.2  & - & 1,148 & 2.8 & 54.8 & 8,780 & 40,310\\
        EEB\&LM-geom~\cite{maksai2019eliminating} & Offline 
    & 22.2  & 27.2& 700 & 3.1 & 61.6 & 5,591 & 41,531\\
    \rowcolor{Gray}
    SORT~\cite{bewley2016simple} & Online  
    & 21.7 & 26.8 & 1,231 & 3.7 & 49.1 & 8,422 & 38,454\\
    EAMTT~\cite{sanchez2016online} & Online 
    & 22.3  & \underline{32.8} & 833 & 5.4 & 52.7 & \underline{7,924} & 38,982\\
    RMOT~\cite{yoon2015bayesian} & Online 
    & 18.6 & 32.6 & \textbf{684} & 5.3 & 53.3 & 12,473 & \underline{36,835}\\
    BiDirLSTM~\cite{yoon2019data} & Online
     & \underline{22.5} & 25.9 & 1,159 & \underline{6.4} & 61.9 & \textbf{7,346} & 39,092\\
    RNN-LSTM~\cite{milan2017online} & Online
     & 19.0 & 17.1& 1,490 & 5.5 & \underline{45.6} & 11,578 & \textbf{36,706} \\
    \midrule
    ArTIST & Online 
    & \textbf{24.2} & \textbf{33.3} & \underline{807} & \textbf{7.1} & \textbf{44.8} & 8,542 & 37,216\\
    \rowcolor{white}
    \midrule
    \multicolumn{9}{c}{All other MOT Methods}\\
    \midrule
    \rowcolor{Gray}
    AMIR15~\cite{sadeghian2017tracking} & Online & 37.6 & 46.0 & 1,026 & 15.8 & 26.8 & 7,933 & 29,397\\
    STRN~\cite{xu2019spatial} & Online & 38.1 & 46.6 & 1,033 & 11.5 & 33.4 & \underline{5,451} & 31,571 \\
    Tracktor++~\cite{bergmann2019tracking} & Online & \underline{44.1} & \underline{46.7} & 1,318 & \underline{18.0} & \textbf{26.2} & 6,477 & \underline{26,577} \\
    DeepMOT-Tracktor~\cite{xu2019deepmot}&Online&\underline{44.1} & 46.0 & 1,347 & 17.2 &\underline{26.6} & 6,085 & 26,917 \\
    FAMNet~\cite{chu2019famnet} & Online & 40.6 & 41.4 & \underline{778} & 12.5 & 34.4  & \textbf{4,678} & 31,018  \\
    \midrule
    ArTIST* & Online & \textbf{45.6} & \textbf{51.0} & \textbf{611} & \textbf{18.9} & 32.5 & 6,334 & \textbf{26,495}\\
    \bottomrule
    \end{tabular}
    \\
    \\
         %%%%%%%%%% MOT16
    \textbf{Results on MOT16 Test Set} \\
    \begin{tabular}{ l c  c c c c c c c }
    \toprule
    Method & Mode  & MOTA $\uparrow$ & IDF1 $\uparrow$& IDs $\downarrow$ & MT $\uparrow$ &  ML $\downarrow$ & FP $\downarrow$ & FN $\downarrow$ \\
    \midrule
    \multicolumn{9}{c}{Geometry-based Tracking Methods}\\
    \midrule
    LP2D~\cite{leal2015motchallenge} & Offline & 35.7 & 34.2 & 915 & 8.7 & 50.7 & 5,084 & 111,163\\
    SMOT~\cite{dicle2013way} & Offline & 29.7 & - & 3,108 & 5.3 & 47.7 & 17,426 & 107,552 \\
    JPDA~\cite{hamid2015joint}&Offline & 26.2 & - & 365 & 4.1 & 67.5 & 3,689 & 130,549 \\
    \rowcolor{Gray}
    EAMTT~\cite{sanchez2016online} & Online & 38.8 & \textbf{42.4} & \textbf{965} & 7.9 & 49.1 & \textbf{8,114} & 102,452  \\
    \midrule
    ArTIST & Online & \textbf{40.0} & 38.9 & 996 & \textbf{11.9} & \textbf{44.5} & 11,500 & \textbf{96,883}\\
     \rowcolor{white}
    \midrule
    \multicolumn{9}{c}{All other MOT Methods}\\
    \midrule
     MHT-BiLSTM~\cite{kim2018multi} & Offline & 42.1 & 47.8 & 753 & 14.9 & 44.4 & 11,637 & 93,172 \\
     MHT-DAM~\cite{kim2015multiple} & Offline & 45.8 & 46.1 & 590 & 16.2 & 43.2 & 6,412 & 91,758 \\
     LMP~\cite{tang2017multiple} & Offline & 48.8 &  51.3 & 481 & 18.2 & 40.1 & 6,654 & 86,245\\
     \rowcolor{Gray}
    AMIR~\cite{sadeghian2017tracking} & Online & 47.2 & 46.3 & 774 & 14.0 & 41.6 & \textbf{2,681} & 92,856 \\
    DMAN~\cite{zhu2018online} & Online & 46.1 & \underline{54.8} & \underline{532} & 17.4 & 42.7 &   7,909 & 89,874 \\
    MOTDT~\cite{long2018real} & Online & 47.6 & 50.9 & 792 & 15.2 & 38.3 & 9,253 & 85,431\\
    Tracktor++~\cite{bergmann2019tracking} & Online & 54.4 & 52.5 & 682 & 19.0 & \underline{36.9} & 3,280 & 79,149 \\
    DeepMOT-Tracktor~\cite{xu2019deepmot} & Online & \underline{54.8} & 53.4 & 645 & \underline{19.1} & 37.0 & \underline{2,955} & \underline{78,765} \\
    STRN~\cite{xu2019spatial} & Online & 48.5 & 53.9 & 747 & 17.0 & \textbf{34.9} & 9,038 & 84,178 \\
       \midrule
    ArTIST*  & Online & \textbf{57.0} & \textbf{59.2} & \textbf{434} & \textbf{21.9} & 37.4 & 3,372 & \textbf{74,679} \\
    \bottomrule
    \end{tabular}
    \\
    \\
    
         %%%%%%% MOT17
         \textbf{Results on MOT2017 Test Set} \\
    \begin{tabular}{ l c  c c c c c c c }
    \toprule
    Method & Mode  & MOTA $\uparrow$ & IDF1 $\uparrow$& IDs $\downarrow$ & MT $\uparrow$ &  ML $\downarrow$ & FP $\downarrow$ & FN $\downarrow$  \\
    \midrule
    \multicolumn{9}{c}{Geometry-based Tracking Methods}\\
    \midrule
    IOU17~\cite{bochinski2017high} & Offline 
    & 45.5  & 39.4 & 5,988 & 15.7 & 40.5 & 19,993 & 281,643 \\
    \rowcolor{Gray}
    SORT17~\cite{bewley2016simple} & Online 
     & 43.1 & 39.8& 4,852 & 12.5 & 42.3 & 28,398 & 287,582 \\
    PHD-GM~\cite{sanchez2019predictor} & Online 
    & \textbf{48.8} & \underline{43.2} & \underline{4,407} & \textbf{19.1} & \textbf{35.2} & \underline{26,260} &\textbf{257,971} \\
    EAMTT~\cite{sanchez2016online} & Online 
     & 42.6 & 41.8& 4,488 & 12.7 & 42.7 & 30,711 & 288,474 \\
    GMPHD-KCF~\cite{kutschbach2017sequential} & Online 
    & 39.6  & 36.6& 5,811 & 8.8 & 43.3 & 50,903 & 284,228\\
    GM-PHD~\cite{eiselein2012real} & Online 
     & 36.4 & 33.9 & 4,607 & 4.1 & 57.3 & \textbf{23,723} & 330,767\\
    \midrule
    ArTIST & Online 
     & \underline{47.4} & \textbf{47.0} & \textbf{3,038} &  \underline{17.6} & \underline{39.5} & 28,726 & \underline{264,945}\\
    %\midrule
    %\multicolumn{10}{c}{RNN-based Tracking Methods (geometry + appearance %and/or social information)}\\
    %\midrule
    \rowcolor{white}
    \midrule
    \multicolumn{9}{c}{All other MOT Methods}\\
    \midrule
    EEB\&LM~\cite{maksai2019eliminating} & Offline 
    & 44.2 & 57.2 & 1,529 & 16.1 & 44.3 & 29,473 & 283,611 \\
    MHT-BiLSTM~\cite{kim2018multi} & Offline 
     & 47.5 & 51.9 & 2,069 & 18.2 & 41.7 & 25,981 & 268,042\\
    MHT-DAM~\cite{kim2015multiple}& Offline & 50.7 & 47.2 & 2,314 & 20.8 & 36.9 & 22,875 & 252,889  \\
    eHAF~\cite{sheng2018heterogeneous}&Offline&51.8 & 54.7 & 1,843 & 23.4 & 37.9 & 33,212 & 236,772 \\
    \rowcolor{Gray}
    MOTDT~\cite{long2018real} & Online & 50.9 & 52.7 & 2,474 & 17.5 & \underline{35.7} & 24,069 & 250,768\\
    DMAN~\cite{zhu2018online} & Online & 48.2 & \underline{55.7} & 2,194 & 19.3 & 38.3 & 26,218 & 263,608 \\
    Tracktor++~\cite{bergmann2019tracking} & Online & 53.5 & 52.3 & 2,072 & 19.5 & 36.6 & \underline{12,201} & 248,047 \\
    DeepMOT-Tracktor~\cite{xu2019deepmot}&Online & \underline{53.7} & 53.8 & \underline{1,947} & 19.4 & 36.6 & \textbf{11,731} & 247,447 \\
    FAMNet~\cite{chu2019famnet} & Online & 52.0 & 48.7 & 3,072 & 19.1 & \textbf{33.4} & 14,138 & 253,616\\
    STRN~\cite{xu2019spatial} & Online & 50.9 & \textbf{56.5} & 2,593 & \underline{20.1} & 37.0 & 27,532 & \underline{246,924} \\ 
    \midrule
    ArTIST*  & Online & \textbf{54.0} & 55.1 & \textbf{1,726} & \textbf{20.7} & 41.0 & 16,549 & \textbf{241,224}\\
    \bottomrule
    \end{tabular}
    \end{tabular}
    }
\end{table}

\subsection{Comparison with the State of the Art}
In this section, we compare our approach with the existing MOT approaches on MOTChallenge and JRDB datasets.

\paragraph{\textbf{Results on MOTChallenge.}}
We compare ArTIST with existing approaches that, as ours, use geometric information from the public detections provided by the benchmarks (top part of each MOTChallenge dataset in Table~\ref{tab:motchallenge}). We also compare ArTIST*, as discussed in Section~\ref{sec:ablation}, with approaches that further utilize appearance information (bottom part of each MOTChallenge dataset in Table~\ref{tab:motchallenge}). For the sake of completeness, we consider both online and offline approaches, however, only online approaches (highlighted rows in Table~\ref{tab:motchallenge}) are directly comparable to ArTIST and ArTIST*. 
The detections provided by these datasets are noisy and of low quality (e.g., ACF detections in 2D MOT2015). This often translates to a considerable number of false/mis-detections which considerably affects the quality of tracking-by-detection approaches. However, as shown in Table~\ref{tab:motchallenge}, despite its simplicity, ArTIST achieves good performance compared to both online and offline geometry-based methods even when they use complicated loss functions~\cite{maksai2019eliminating} or approximation methods~\cite{hamid2015joint}.
Importantly, despite the fact that our approach does not rely on the MOTChallenge training data, as opposed to some baselines such as~\cite{maksai2019eliminating,milan2017online}, it outperforms these methods by a considerable margin. Note that the good performance of PHD-GM~\cite{sanchez2019predictor} in Table~\ref{tab:motchallenge} (MOT2017), which is a motion prediction model, is due to its use of additional hand-crafted features, e.g., the good-features-to-track~\cite{shi1994good} and pyramids of Lucas-Kanade optical flow~\cite{bouguet2001pyramidal}, to estimate the camera/background motion. 

When comparing ArTIST* with state-of-the-art appearance-based tracking methods, we observe that the methods that follow the refinement approach of Tracktor~\cite{bergmann2019tracking,xu2019deepmot}, including ours, achieve better performance than other techniques.
Note that almost all state-of-the-art methods in Table~\ref{tab:motchallenge} utilize a form of person re-identification to better preserve the tracklet identities over a sequence. However, using such modules requires additional training and was shown to be highly dependent on the target domain~\cite{fu2019self,deng2018image,fan2018unsupervised}. Thanks to our probabilistic geometry-based scoring function and ArTIST's capability to inpaint missing detections, ArTIST* achieves superior performance in preserving the tracklet identities without an additional person re-identification module and additional training, as evidenced by the much lower ID switches, higher MOTA, IDF1, and MT. Note that the low FN of ArTIST* validates the fact that the inpainted bounding boxes encode natural and valid motions.

\begin{table*}[t]
    \centering
    \caption{\textbf{Results on the JRDB test set.} Note that ArTIST is only trained on PathTrack, as in our other experiments.}
    \label{tab:JRDB}
    \scriptsize
    \tabcolsep=0.1cm
    \begin{tabular}{ l c c c c }
    \toprule
    Method & MOTA $\uparrow$ & IDs $\downarrow$ & FP $\downarrow$ & FN $\downarrow$ \\
    \midrule
    DeepSORT~\cite{Wojke2017simple} & \underline{0.232} & \textbf{5,296} & 78,947 & \underline{650,478}\\
    JRMOT2D~\cite{shenoi2020jrmot}  & 0.225 & 7,719 & \textbf{65,550} & 667,783\\
    % \midrule
    ArTIST & \textbf{0.236} & \underline{6,684} & \underline{70,557} &  \underline{654,083}\\
    \bottomrule
    \end{tabular}
\end{table*}

\paragraph{\textbf{Results on JRDB.}} 
We also evaluate our approach on the recent JRDB 2D Tracking Challenge~\cite{martin2019jrdb}. This dataset contains very challenging scenarios depicting highly crowded scenes with drastically different viewpoints (from a robot's view). For this experiment, we use the detections provided by the challenge.
Table~\ref{tab:JRDB} compares ArTIST with existing methods that are publicly available on the challenge leaderboard\footnote{Available at \texttt{https://jrdb.stanford.edu/leaderboards/2d\_tracking}.}. 
Both DeepSort~\cite{Wojke2017simple} and JRMOT2D~\cite{shenoi2020jrmot} leverage the appearance information for tracking. Furthermore, JRMOT2D builds on the Aligned-ReID~\cite{zhang2017alignedreid} framework, which was trained on the JRDB training set. Unlike these methods, ArTIST, which yields the best MOTA performance, is only trained on PathTrack and only relies on geometric information for tracking.

\section{Conclusion}
We have introduced an online MOT framework based on a probabilistic autoregressive generative model of natural motions. Specifically, we have employed this model to both score tracklets for detection assignment purposes and inpaint tracklets to account for missing detections. 
Our results on the MOTChallenge benchmark and JRDB dataset have shown the benefits of relying on a probabilistic representation of motion. Notably, without being trained specifically for these benchmarks, our framework yields state-of-the-art performance. 

\clearpage
% ---- Bibliography ----
\bibliographystyle{splncs04}
\bibliography{egbib}

\begin{thebibliography}{10}
\providecommand{\url}[1]{\texttt{#1}}
\providecommand{\urlprefix}{URL }
\providecommand{\doi}[1]{https://doi.org/#1}

\bibitem{alahi2016social}
Alahi, A., Goel, K., Ramanathan, V., Robicquet, A., Fei-Fei, L., Savarese, S.:
  Social lstm: Human trajectory prediction in crowded spaces. In: Proceedings
  of the IEEE conference on computer vision and pattern recognition. pp.
  961--971 (2016)

\bibitem{aliakbarian2019learning}
Aliakbarian, M.S., Saleh, F.S., Salzmann, M., Petersson, L., Gould, S.,
  Habibian, A.: Learning variations in human motion via mix-and-match
  perturbation. arXiv preprint arXiv:1908.00733  (2019)

\bibitem{aliakbarian2019sampling}
Aliakbarian, S., Saleh, F.S., Salzmann, M., Petersson, L., Gould, S.: Sampling
  good latent variables via cpp-vaes: Vaes with condition posterior as prior.
  arXiv preprint arXiv:1912.08521  (2019)

\bibitem{aliakbarian2020stochastic}
Aliakbarian, S., Saleh, F.S., Salzmann, M., Petersson, L., Gould, S.: A
  stochastic conditioning scheme for diverse human motion prediction. In:
  Proceedings of the IEEE international conference on computer vision (2020)

\bibitem{andriluka2008people}
Andriluka, M., Roth, S., Schiele, B.: People-tracking-by-detection and
  people-detection-by-tracking. In: 2008 IEEE Conference on computer vision and
  pattern recognition. pp.~1--8. IEEE (2008)

\bibitem{barsoum2018hp}
Barsoum, E., Kender, J., Liu, Z.: Hp-gan: Probabilistic 3d human motion
  prediction via gan. In: Proceedings of the IEEE Conference on Computer Vision
  and Pattern Recognition Workshops. pp. 1418--1427 (2018)

\bibitem{bergmann2019tracking}
Bergmann, P., Meinhardt, T., Leal-Taixe, L.: Tracking without bells and
  whistles. arXiv preprint arXiv:1903.05625  (2019)

\bibitem{bernardin2008evaluating}
Bernardin, K., Stiefelhagen, R.: Evaluating multiple object tracking
  performance: the clear mot metrics. Journal on Image and Video Processing
  \textbf{2008}, ~1 (2008)

\bibitem{bertinetto2016staple}
Bertinetto, L., Valmadre, J., Golodetz, S., Miksik, O., Torr, P.H.: Staple:
  Complementary learners for real-time tracking. In: Proceedings of the IEEE
  conference on computer vision and pattern recognition. pp. 1401--1409 (2016)

\bibitem{bewley2016simple}
Bewley, A., Ge, Z., Ott, L., Ramos, F., Upcroft, B.: Simple online and realtime
  tracking. In: 2016 IEEE International Conference on Image Processing (ICIP).
  pp. 3464--3468. IEEE (2016)

\bibitem{blackman2004multiple}
Blackman, S.S.: Multiple hypothesis tracking for multiple target tracking. IEEE
  Aerospace and Electronic Systems Magazine  \textbf{19}(1),  5--18 (2004)

\bibitem{bochinski2017high}
Bochinski, E., Eiselein, V., Sikora, T.: High-speed tracking-by-detection
  without using image information. In: 2017 14th IEEE International Conference
  on Advanced Video and Signal Based Surveillance (AVSS). pp.~1--6. IEEE (2017)

\bibitem{bouguet2001pyramidal}
Bouguet, J.Y., et~al.: Pyramidal implementation of the affine lucas kanade
  feature tracker description of the algorithm. Intel Corporation
  \textbf{5}(1-10), ~4 (2001)

\bibitem{breitenstein2010online}
Breitenstein, M.D., Reichlin, F., Leibe, B., Koller-Meier, E., Van~Gool, L.:
  Online multiperson tracking-by-detection from a single, uncalibrated camera.
  IEEE transactions on pattern analysis and machine intelligence
  \textbf{33}(9),  1820--1833 (2010)

\bibitem{butepage2017deep}
Butepage, J., Black, M.J., Kragic, D., Kjellstrom, H.: Deep representation
  learning for human motion prediction and classification. In: Proceedings of
  the IEEE conference on computer vision and pattern recognition. pp.
  6158--6166 (2017)

\bibitem{chen2017enhancing}
Chen, J., Sheng, H., Zhang, Y., Xiong, Z.: Enhancing detection model for
  multiple hypothesis tracking. In: Proceedings of the IEEE Conference on
  Computer Vision and Pattern Recognition Workshops. pp. 18--27 (2017)

\bibitem{chen2017online}
Chen, L., Ai, H., Shang, C., Zhuang, Z., Bai, B.: Online multi-object tracking
  with convolutional neural networks. In: 2017 IEEE International Conference on
  Image Processing (ICIP). pp. 645--649. IEEE (2017)

\bibitem{choi2015near}
Choi, W.: Near-online multi-target tracking with aggregated local flow
  descriptor. In: Proceedings of the IEEE international conference on computer
  vision. pp. 3029--3037 (2015)

\bibitem{chu2019online}
Chu, P., Fan, H., Tan, C.C., Ling, H.: Online multi-object tracking with
  instance-aware tracker and dynamic model refreshment. In: 2019 IEEE Winter
  Conference on Applications of Computer Vision (WACV). pp. 161--170. IEEE
  (2019)

\bibitem{chu2019famnet}
Chu, P., Ling, H.: Famnet: Joint learning of feature, affinity and
  multi-dimensional assignment for online multiple object tracking. arXiv
  preprint arXiv:1904.04989  (2019)

\bibitem{chu2017online}
Chu, Q., Ouyang, W., Li, H., Wang, X., Liu, B., Yu, N.: Online multi-object
  tracking using cnn-based single object tracker with spatial-temporal
  attention mechanism. In: Proceedings of the IEEE International Conference on
  Computer Vision. pp. 4836--4845 (2017)

\bibitem{deng2018image}
Deng, W., Zheng, L., Ye, Q., Kang, G., Yang, Y., Jiao, J.: Image-image domain
  adaptation with preserved self-similarity and domain-dissimilarity for person
  re-identification. In: Proceedings of the IEEE conference on computer vision
  and pattern recognition. pp. 994--1003 (2018)

\bibitem{dicle2013way}
Dicle, C., Camps, O.I., Sznaier, M.: The way they move: Tracking multiple
  targets with similar appearance. In: Proceedings of the IEEE international
  conference on computer vision. pp. 2304--2311 (2013)

\bibitem{dollar2014fast}
Doll{\'a}r, P., Appel, R., Belongie, S., Perona, P.: Fast feature pyramids for
  object detection. IEEE transactions on pattern analysis and machine
  intelligence  \textbf{36}(8),  1532--1545 (2014)

\bibitem{eiselein2012real}
Eiselein, V., Arp, D., P{\"a}tzold, M., Sikora, T.: Real-time multi-human
  tracking using a probability hypothesis density filter and multiple
  detectors. In: 2012 IEEE Ninth International Conference on Advanced Video and
  Signal-Based Surveillance. pp. 325--330. IEEE (2012)

\bibitem{fan2018unsupervised}
Fan, H., Zheng, L., Yan, C., Yang, Y.: Unsupervised person re-identification:
  Clustering and fine-tuning. ACM Transactions on Multimedia Computing,
  Communications, and Applications (TOMM)  \textbf{14}(4),  1--18 (2018)

\bibitem{fang2018recurrent}
Fang, K., Xiang, Y., Li, X., Savarese, S.: Recurrent autoregressive networks
  for online multi-object tracking. In: 2018 IEEE Winter Conference on
  Applications of Computer Vision (WACV). pp. 466--475. IEEE (2018)

\bibitem{felzenszwalb2009object}
Felzenszwalb, P.F., Girshick, R.B., McAllester, D., Ramanan, D.: Object
  detection with discriminatively trained part-based models. IEEE transactions
  on pattern analysis and machine intelligence  \textbf{32}(9),  1627--1645
  (2009)

\bibitem{fu2019self}
Fu, Y., Wei, Y., Wang, G., Zhou, Y., Shi, H., Huang, T.S.: Self-similarity
  grouping: A simple unsupervised cross domain adaptation approach for person
  re-identification. In: Proceedings of the IEEE International Conference on
  Computer Vision. pp. 6112--6121 (2019)

\bibitem{girshick2015fast}
Girshick, R.: Fast r-cnn. In: Proceedings of the IEEE international conference
  on computer vision. pp. 1440--1448 (2015)

\bibitem{gui2018adversarial}
Gui, L.Y., Wang, Y.X., Liang, X., Moura, J.M.: Adversarial geometry-aware human
  motion prediction. In: Proceedings of the European Conference on Computer
  Vision (ECCV). pp. 786--803 (2018)

\bibitem{gupta2018social}
Gupta, A., Johnson, J., Fei-Fei, L., Savarese, S., Alahi, A.: Social gan:
  Socially acceptable trajectories with generative adversarial networks. In:
  Proceedings of the IEEE Conference on Computer Vision and Pattern
  Recognition. pp. 2255--2264 (2018)

\bibitem{hamid2015joint}
Hamid~Rezatofighi, S., Milan, A., Zhang, Z., Shi, Q., Dick, A., Reid, I.: Joint
  probabilistic data association revisited. In: Proceedings of the IEEE
  international conference on computer vision. pp. 3047--3055 (2015)

\bibitem{he2016deep}
He, K., Zhang, X., Ren, S., Sun, J.: Deep residual learning for image
  recognition. In: Proceedings of the IEEE conference on computer vision and
  pattern recognition. pp. 770--778 (2016)

\bibitem{hermans2017defense}
Hermans, A., Beyer, L., Leibe, B.: In defense of the triplet loss for person
  re-identification. arXiv preprint arXiv:1703.07737  (2017)

\bibitem{hong2016online}
Hong~Yoon, J., Lee, C.R., Yang, M.H., Yoon, K.J.: Online multi-object tracking
  via structural constraint event aggregation. In: Proceedings of the IEEE
  Conference on computer vision and pattern recognition. pp. 1392--1400 (2016)

\bibitem{ionescu2013human3}
Ionescu, C., Papava, D., Olaru, V., Sminchisescu, C.: Human3. 6m: Large scale
  datasets and predictive methods for 3d human sensing in natural environments.
  IEEE transactions on pattern analysis and machine intelligence
  \textbf{36}(7),  1325--1339 (2013)

\bibitem{kim2015multiple}
Kim, C., Li, F., Ciptadi, A., Rehg, J.M.: Multiple hypothesis tracking
  revisited. In: Proceedings of the IEEE International Conference on Computer
  Vision. pp. 4696--4704 (2015)

\bibitem{kim2018multi}
Kim, C., Li, F., Rehg, J.M.: Multi-object tracking with neural gating using
  bilinear lstm. In: Proceedings of the European Conference on Computer Vision
  (ECCV). pp. 200--215 (2018)

\bibitem{kingma2014adam}
Kingma, D.P., Ba, J.: Adam: A method for stochastic optimization. arXiv
  preprint arXiv:1412.6980  (2014)

\bibitem{kundu2018bihmp}
Kundu, J.N., Gor, M., Babu, R.V.: Bihmp-gan: Bidirectional 3d human motion
  prediction gan. arXiv preprint arXiv:1812.02591  (2018)

\bibitem{kutschbach2017sequential}
Kutschbach, T., Bochinski, E., Eiselein, V., Sikora, T.: Sequential sensor
  fusion combining probability hypothesis density and kernelized correlation
  filters for multi-object tracking in video data. In: 2017 14th IEEE
  International Conference on Advanced Video and Signal Based Surveillance
  (AVSS). pp.~1--5. IEEE (2017)

\bibitem{leal2016learning}
Leal-Taix{\'e}, L., Canton-Ferrer, C., Schindler, K.: Learning by tracking:
  Siamese cnn for robust target association. In: Proceedings of the IEEE
  Conference on Computer Vision and Pattern Recognition Workshops. pp. 33--40
  (2016)

\bibitem{leal2015motchallenge}
Leal-Taix{\'e}, L., Milan, A., Reid, I., Roth, S., Schindler, K.: Motchallenge
  2015: Towards a benchmark for multi-target tracking. arXiv preprint
  arXiv:1504.01942  (2015)

\bibitem{leal2017tracking}
Leal-Taix{\'e}, L., Milan, A., Schindler, K., Cremers, D., Reid, I., Roth, S.:
  Tracking the trackers: an analysis of the state of the art in multiple object
  tracking. arXiv preprint arXiv:1704.02781  (2017)

\bibitem{li2018convolutional}
Li, C., Zhang, Z., Sun~Lee, W., Hee~Lee, G.: Convolutional sequence to sequence
  model for human dynamics. In: Proceedings of the IEEE Conference on Computer
  Vision and Pattern Recognition. pp. 5226--5234 (2018)

\bibitem{liang2018lstm}
Liang, Y., Zhou, Y.: Lstm multiple object tracker combining multiple cues. In:
  2018 25th IEEE International Conference on Image Processing (ICIP). pp.
  2351--2355. IEEE (2018)

\bibitem{lin2017feature}
Lin, T.Y., Doll{\'a}r, P., Girshick, R., He, K., Hariharan, B., Belongie, S.:
  Feature pyramid networks for object detection. In: Proceedings of the IEEE
  conference on computer vision and pattern recognition. pp. 2117--2125 (2017)

\bibitem{lin2018human}
Lin, X., Amer, M.R.: Human motion modeling using dvgans. arXiv preprint
  arXiv:1804.10652  (2018)

\bibitem{long2018real}
Long, C., Haizhou, A., Zijie, Z., Chong, S.: Real-time multiple people tracking
  with deeply learned candidate selection and person re-identification. In:
  ICME. vol.~5, p.~8 (2018)

\bibitem{maksai2019eliminating}
Maksai, A., Fua, P.: Eliminating exposure bias and metric mismatch in multiple
  object tracking. In: Proceedings of the IEEE Conference on Computer Vision
  and Pattern Recognition. pp. 4639--4648 (2019)

\bibitem{maksai2017non}
Maksai, A., Wang, X., Fleuret, F., Fua, P.: Non-markovian globally consistent
  multi-object tracking. In: Proceedings of the IEEE International Conference
  on Computer Vision. pp. 2544--2554 (2017)

\bibitem{manen2017pathtrack}
Manen, S., Gygli, M., Dai, D., Van~Gool, L.: Pathtrack: Fast trajectory
  annotation with path supervision. In: Proceedings of the IEEE International
  Conference on Computer Vision. pp. 290--299 (2017)

\bibitem{mao2019learning}
Mao, W., Liu, M., Salzmann, M., Li, H.: Learning trajectory dependencies for
  human motion prediction. arXiv preprint arXiv:1908.05436  (2019)

\bibitem{martin2019jrdb}
Mart{\'\i}n-Mart{\'\i}n, R., Rezatofighi, H., Shenoi, A., Patel, M., Gwak, J.,
  Dass, N., Federman, A., Goebel, P., Savarese, S.: Jrdb: A dataset and
  benchmark for visual perception for navigation in human environments. arXiv
  preprint arXiv:1910.11792  (2019)

\bibitem{martinez2017human}
Martinez, J., Black, M.J., Romero, J.: On human motion prediction using
  recurrent neural networks. In: 2017 IEEE Conference on Computer Vision and
  Pattern Recognition (CVPR). pp. 4674--4683. IEEE (2017)

\bibitem{milan2016mot16}
Milan, A., Leal-Taix{\'e}, L., Reid, I., Roth, S., Schindler, K.: Mot16: A
  benchmark for multi-object tracking. arXiv preprint arXiv:1603.00831  (2016)

\bibitem{milan2017online}
Milan, A., Rezatofighi, S.H., Dick, A., Reid, I., Schindler, K.: Online
  multi-target tracking using recurrent neural networks. In: Thirty-First AAAI
  Conference on Artificial Intelligence (2017)

\bibitem{milan2013continuous}
Milan, A., Roth, S., Schindler, K.: Continuous energy minimization for
  multitarget tracking. IEEE transactions on pattern analysis and machine
  intelligence  \textbf{36}(1),  58--72 (2013)

\bibitem{munkres1957algorithms}
Munkres, J.: Algorithms for the assignment and transportation problems. Journal
  of the society for industrial and applied mathematics  \textbf{5}(1),  32--38
  (1957)

\bibitem{oord2016pixel}
Oord, A.v.d., Kalchbrenner, N., Kavukcuoglu, K.: Pixel recurrent neural
  networks. arXiv preprint arXiv:1601.06759  (2016)

\bibitem{pascanu2013difficulty}
Pascanu, R., Mikolov, T., Bengio, Y.: On the difficulty of training recurrent
  neural networks. In: International conference on machine learning. pp.
  1310--1318 (2013)

\bibitem{paszke2017automatic}
Paszke, A., Gross, S., Chintala, S., Chanan, G., Yang, E., DeVito, Z., Lin, Z.,
  Desmaison, A., Antiga, L., Lerer, A.: Automatic differentiation in {PyTorch}.
  In: NIPS Autodiff Workshop (2017)

\bibitem{pavllo2019modeling}
Pavllo, D., Feichtenhofer, C., Auli, M., Grangier, D.: Modeling human motion
  with quaternion-based neural networks. arXiv preprint arXiv:1901.07677
  (2019)

\bibitem{pavllo2018quaternet}
Pavllo, D., Grangier, D., Auli, M.: Quaternet: A quaternion-based recurrent
  model for human motion. arXiv preprint arXiv:1805.06485  (2018)

\bibitem{pirsiavash2011globally}
Pirsiavash, H., Ramanan, D., Fowlkes, C.C.: Globally-optimal greedy algorithms
  for tracking a variable number of objects. In: CVPR 2011. pp. 1201--1208.
  IEEE (2011)

\bibitem{ran2019robust}
Ran, N., Kong, L., Wang, Y., Liu, Q.: A robust multi-athlete tracking algorithm
  by exploiting discriminant features and long-term dependencies. In:
  International Conference on Multimedia Modeling. pp. 411--423. Springer
  (2019)

\bibitem{ristani2016performance}
Ristani, E., Solera, F., Zou, R., Cucchiara, R., Tomasi, C.: Performance
  measures and a data set for multi-target, multi-camera tracking. In: European
  Conference on Computer Vision. pp. 17--35. Springer (2016)

\bibitem{ristani2018features}
Ristani, E., Tomasi, C.: Features for multi-target multi-camera tracking and
  re-identification. In: Proceedings of the IEEE conference on computer vision
  and pattern recognition. pp. 6036--6046 (2018)

\bibitem{sadeghian2017tracking}
Sadeghian, A., Alahi, A., Savarese, S.: Tracking the untrackable: Learning to
  track multiple cues with long-term dependencies. In: Proceedings of the IEEE
  International Conference on Computer Vision. pp. 300--311 (2017)

\bibitem{sanchez2019predictor}
Sanchez-Matilla, R., Cavallaro, A.: A predictor of moving objects for
  first-person vision. In: 2019 IEEE International Conference on Image
  Processing (ICIP). pp. 2189--2193. IEEE (2019)

\bibitem{sanchez2016online}
Sanchez-Matilla, R., Poiesi, F., Cavallaro, A.: Online multi-target tracking
  with strong and weak detections. In: European Conference on Computer Vision.
  pp. 84--99. Springer (2016)

\bibitem{sheng2018heterogeneous}
Sheng, H., Zhang, Y., Chen, J., Xiong, Z., Zhang, J.: Heterogeneous association
  graph fusion for target association in multiple object tracking. IEEE
  Transactions on Circuits and Systems for Video Technology  \textbf{29}(11),
  3269--3280 (2018)

\bibitem{shenoi2020jrmot}
Shenoi, A., Patel, M., Gwak, J., Goebel, P., Sadeghian, A., Rezatofighi, H.,
  Martin-Martin, R., Savarese, S.: Jrmot: A real-time 3d multi-object tracker
  and a new large-scale dataset. arXiv preprint arXiv:2002.08397  (2020)

\bibitem{shi1994good}
Shi, J., et~al.: Good features to track. In: 1994 Proceedings of IEEE
  conference on computer vision and pattern recognition. pp. 593--600. IEEE
  (1994)

\bibitem{son2017multi}
Son, J., Baek, M., Cho, M., Han, B.: Multi-object tracking with quadruplet
  convolutional neural networks. In: Proceedings of the IEEE conference on
  computer vision and pattern recognition. pp. 5620--5629 (2017)

\bibitem{song2018vital}
Song, Y., Ma, C., Wu, X., Gong, L., Bao, L., Zuo, W., Shen, C., Lau, R.W.,
  Yang, M.H.: Vital: Visual tracking via adversarial learning. In: Proceedings
  of the IEEE Conference on Computer Vision and Pattern Recognition. pp.
  8990--8999 (2018)

\bibitem{sun2019deep}
Sun, S., Akhtar, N., Song, H., Mian, A.S., Shah, M.: Deep affinity network for
  multiple object tracking. IEEE transactions on pattern analysis and machine
  intelligence  (2019)

\bibitem{tang2015subgraph}
Tang, S., Andres, B., Andriluka, M., Schiele, B.: Subgraph decomposition for
  multi-target tracking. In: Proceedings of the IEEE Conference on Computer
  Vision and Pattern Recognition. pp. 5033--5041 (2015)

\bibitem{tang2017multiple}
Tang, S., Andriluka, M., Andres, B., Schiele, B.: Multiple people tracking by
  lifted multicut and person re-identification. In: Proceedings of the IEEE
  Conference on Computer Vision and Pattern Recognition. pp. 3539--3548 (2017)

\bibitem{walker2017pose}
Walker, J., Marino, K., Gupta, A., Hebert, M.: The pose knows: Video
  forecasting by generating pose futures. In: Computer Vision (ICCV), 2017 IEEE
  International Conference on. pp. 3352--3361. IEEE (2017)

\bibitem{wan2018online}
Wan, X., Wang, J., Zhou, S.: An online and flexible multi-object tracking
  framework using long short-term memory. In: Proceedings of the IEEE
  Conference on Computer Vision and Pattern Recognition Workshops. pp.
  1230--1238 (2018)

\bibitem{williams1989learning}
Williams, R.J., Zipser, D.: A learning algorithm for continually running fully
  recurrent neural networks. Neural computation  \textbf{1}(2),  270--280
  (1989)

\bibitem{Wojke2017simple}
Wojke, N., Bewley, A., Paulus, D.: Simple online and realtime tracking with a
  deep association metric. In: 2017 IEEE International Conference on Image
  Processing (ICIP). pp. 3645--3649. IEEE (2017).
  \doi{10.1109/ICIP.2017.8296962}

\bibitem{xu2019spatial}
Xu, J., Cao, Y., Zhang, Z., Hu, H.: Spatial-temporal relation networks for
  multi-object tracking. arXiv preprint arXiv:1904.11489  (2019)

\bibitem{xu2019deepmot}
Xu, Y., Ban, Y., Alameda-Pineda, X., Horaud, R.: Deepmot: A differentiable
  framework for training multiple object trackers. arXiv preprint
  arXiv:1906.06618  (2019)

\bibitem{yan2018mt}
Yan, X., Rastogi, A., Villegas, R., Sunkavalli, K., Shechtman, E., Hadap, S.,
  Yumer, E., Lee, H.: Mt-vae: Learning motion transformations to generate
  multimodal human dynamics. In: European Conference on Computer Vision. pp.
  276--293. Springer (2018)

\bibitem{yang2012online}
Yang, B., Nevatia, R.: Online learned discriminative part-based appearance
  models for multi-human tracking. In: European Conference on Computer Vision.
  pp. 484--498. Springer (2012)

\bibitem{yang2016exploit}
Yang, F., Choi, W., Lin, Y.: Exploit all the layers: Fast and accurate cnn
  object detector with scale dependent pooling and cascaded rejection
  classifiers. In: Proceedings of the IEEE conference on computer vision and
  pattern recognition. pp. 2129--2137 (2016)

\bibitem{yoon2015bayesian}
Yoon, J.H., Yang, M.H., Lim, J., Yoon, K.J.: Bayesian multi-object tracking
  using motion context from multiple objects. In: 2015 IEEE Winter Conference
  on Applications of Computer Vision. pp. 33--40. IEEE (2015)

\bibitem{yoon2019data}
Yoon, K., Kim, D.Y., Yoon, Y.C., Jeon, M.: Data association for multi-object
  tracking via deep neural networks. Sensors  \textbf{19}(3), ~559 (2019)

\bibitem{yoon2018multiple}
Yoon, K., Song, Y.m., Jeon, M.: Multiple hypothesis tracking algorithm for
  multi-target multi-camera tracking with disjoint views. IET Image Processing
  \textbf{12}(7),  1175--1184 (2018)

\bibitem{yoon2018online}
Yoon, Y.c., Boragule, A., Song, Y.m., Yoon, K., Jeon, M.: Online multi-object
  tracking with historical appearance matching and scene adaptive detection
  filtering. In: 2018 15th IEEE International conference on advanced video and
  signal based surveillance (AVSS). pp.~1--6. IEEE (2018)

\bibitem{zhai2018deep}
Zhai, M., Chen, L., Mori, G., Javan~Roshtkhari, M.: Deep learning of appearance
  models for online object tracking. In: Proceedings of the European Conference
  on Computer Vision (ECCV). pp.~0--0 (2018)

\bibitem{zhang2013structure}
Zhang, L., van~der Maaten, L.: Structure preserving object tracking. In:
  Proceedings of the IEEE conference on computer vision and pattern
  recognition. pp. 1838--1845 (2013)

\bibitem{zhang2018learning}
Zhang, T., Xu, C., Yang, M.H.: Learning multi-task correlation particle filters
  for visual tracking. IEEE transactions on pattern analysis and machine
  intelligence  \textbf{41}(2),  365--378 (2018)

\bibitem{zhang2018robust}
Zhang, T., Xu, C., Yang, M.H.: Robust structural sparse tracking. IEEE
  transactions on pattern analysis and machine intelligence  \textbf{41}(2),
  473--486 (2018)

\bibitem{zhang2017alignedreid}
Zhang, X., Luo, H., Fan, X., Xiang, W., Sun, Y., Xiao, Q., Jiang, W., Zhang,
  C., Sun, J.: Alignedreid: Surpassing human-level performance in person
  re-identification. arXiv preprint arXiv:1711.08184  (2017)

\bibitem{zhu2018online}
Zhu, J., Yang, H., Liu, N., Kim, M., Zhang, W., Yang, M.H.: Online multi-object
  tracking with dual matching attention networks. In: Proceedings of the
  European Conference on Computer Vision (ECCV). pp. 366--382 (2018)

\end{thebibliography}

\clearpage

\appendix
\section{Human Motion Prediction: Related Work}
Due to the success of RNN-based methods at modeling sequence-to-sequence learning problems, many attempts have been made to address motion prediction within a recurrent framework~\cite{martinez2017human,gui2018adversarial,walker2017pose,kundu2018bihmp,barsoum2018hp,pavllo2019modeling,pavllo2018quaternet,aliakbarian2020stochastic}. Typically, these approaches try to learn a mapping from the observed sequence of poses to the future sequence. Another group of study addresses this problem using feed-forward models~\cite{mao2019learning,li2018convolutional,butepage2017deep}, either with fully-connected~\cite{butepage2017deep}, convolutional~\cite{li2018convolutional}, or more recently, graph neural networks~\cite{mao2019learning}. While deterministic approaches tend to produce accurate predictions, a number of studies utilize generative models~\cite{barsoum2018hp,kundu2018bihmp,aliakbarian2019sampling,lin2018human,yan2018mt,aliakbarian2019learning} for the task of human motion prediction.  In contrast to aforementioned approaches, which either cannot directly measure the likelihood of motions~\cite{barsoum2018hp,kundu2018bihmp} or can only roughly approximate it~\cite{aliakbarian2019sampling,yan2018mt,aliakbarian2019learning}, ArTIST can explicitly compute the likelihood of the generated motions. This is an import feature of our model that helps finding a very likely and natural motion without having access to the ground-truth motion or an oracle.

\section{Additional Information on Tracklet Rejection}
\begin{figure*}[t]
    \centering
    \includegraphics[width=0.9\textwidth]{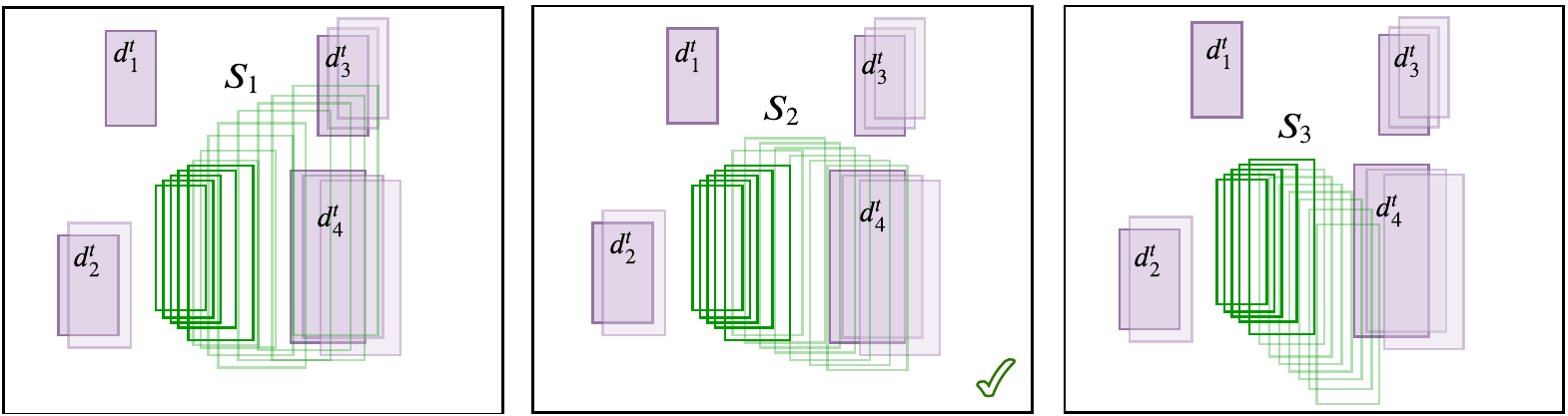}
    \caption{\textbf{Inpainting with ArTIST.} Let us consider three plausible sampled continuations (transparent boxes) of the green tracklet, which was last observed at time $t-\delta_{gap}$. The solid boxes $d_1^t$ to $d_4^t$ are the detections at the current time $t$. We also show the detections for the next $t_{TRS}$ time-steps (if available) as transparent purple boxes. Given observations (the solid green boxes), ArTIST inpaints the gap of length $\delta_{gap}$. From a geometric point of view, all of these inpainted candidates are valid and  are highly likely to be natural continuations of the observed tracklet because we sample them. However, only one of them, if any, can be valid. For each inpainted candidate, if the last sampled bounding box (at time $t$) does not have a considerable overlap with one of the detections at time $t$, the candidate will be rejected (e.g., the case of $S_3$). Otherwise, we continue inpainting for an additional $t_{TRS}$ frames. Then, the tracklet with maximum sum of IOUs with the detections from $t$ to $t_{TRS}$ will be selected (e.g., the case of $S_2$). The likelihood of the selected tracklet will then be used in the assignment step. If the selected candidate is assigned to this tracklet after the assignment step, we update the tracklet with the observations, the inpainted part, and the detection at time $t$ to form a complete trajectory.}
    \label{fig:diverse_sampling}
\end{figure*}

As discussed in the main paper, our model is capable of inpainting the missing observations when a detector fails to detect an object for a few frames. Our model also accounts for the stochasticity of human motion, and thus generates multiple plausible candidates for inpainting. To select one of these candidates, if there is one to be selected, we compute the intersection over union (IOU) of the last generated bounding box with all the detections in the scene. The model then selects the candidate with the highest IOU, if it surpasses a threshold. However, in some cases, the last generated bounding box of one of the candidates may overlap with a false detection or a detection for another object (belonging to a different tracklet). As illustrated in Fig.~\ref{fig:diverse_sampling}, to account for these ambiguities, we continue predicting  boxes for all candidates for $t_{TRS}$ frames. We then compute the IOU  with the detections of not only the current frame, but also the $t_{TRS}$ frames ahead. ArTIST then selects the candidate with the maximum sum of IOUs. This allows us to ignore candidates matching a false detection or a detection for another object moving in a different direction. However, this may not be enough to disambiguate all cases, e.g., the detections belonging to other tracklets that are close-by and moving in the same direction. We consider some of these cases in our assignment, discussed in Section 3.2 of the main paper. Note that, in practice, we use a small $t_{TRS}$, e.g., 2 or 3 frames, and thus our approach can still be considered as online tracking.
{}

\section{Implementation Details}
The architecture of our model consists of a single LSTM layer with 512 hidden units. It takes as input a 4D motion velocity representation, passes it through a fully-connected layer with 512 hidden units followed by ReLU non-linearity, and produces a residual 4D vector, which is added to the input to generate the final representation. To map the output of the LSTM to a probability distribution for each component of the motion velocity, we use 4 fully-connected layers followed by softmax activations, resulting in a $4\times K$ representation, where $K=1024$ is the number of clusters. We train our model on a single GPU with the Adam optimizer~\cite{kingma2014adam} for 150K iterations. We use a learning rate of 0.001 and a mini-batch size of 256. To avoid exploding gradients, we use the gradient-clipping technique of~\cite{pascanu2013difficulty} for all layers in the network. Since we use the ground-truth boxes during training, we apply random jitter to the boxes to simulate the noise produced by a detector. We train our model with sequences of arbitrary length in each mini-batch. During training, we use the teacher forcing technique of~\cite{williams1989learning}, in which ArTIST chooses with probability $P_{tf}$ whether to use its own output (a sampled bounding box) at the previous time-step or the ground-truth bounding box to compute the velocity at each time-step. We use $P_{tf}=0.2$ for the frames occurring after 70\% of the sequence length.
For our online tracking pipeline, we terminate a tracklet if it has not been observed for 60 frames. For tracklet rejection in the case of inpainting, we use an IOU threshold of 0.5 and set $t_{TRS}=2$ for low frame-rate videos and $t_{TRS}=3$ for high frame-rate ones. During multinomial sampling, we sample $\mathcal{S} = 30$ candidate tracklets. 
For the human motion prediction prediction, we use the same architecture. The only difference is the input/output representation. Human poses are represented in 3D position space, with 32 joins in $xyz$ space. Thus, the the input to ArTIST is a 96-dimensional representation for motion velocity. As also stated in the main paper,  unlike tracking where we assume independence between bounding box parameters, here we
consider all joints to be related to each other. Therefore, we cluster the entire
pose velocities into 1024 clusters.
We implemented our model using the Pytorch framework of~\cite{paszke2017automatic}.

\section{Evaluation Metrics}
Several metrics are commonly used to evaluate the quality of a tracking system~\cite{ristani2016performance,bernardin2008evaluating}. The main one is MOTA, which combines three error sources: false positives, false negatives and identity switches. A higher MOTA score implies better performance. Another important metric is IDF1, i.e., the ratio of correctly identified detections over the average number of ground-truth and computed detections. The number of identity switches, IDs, is also frequently reported. Furthermore, the following metrics provide finer details on the performance of a tracking system: mostly tracked (MT) and mostly lost (ML), that are respectively the ratio of ground-truth trajectories that are covered/lost by the tracker for at least 80\% of their respective life span; False positives (FP) and false negatives (FN). 
All  metrics were computed using the official evaluation code provided by the MOTChallenge benchmark.

\end{document}